\documentclass[11pt, a4paper]{article}

\usepackage[utf8]{inputenc}
\usepackage[T1]{fontenc}
\usepackage{geometry}
\usepackage{times}
\usepackage{amsmath, amssymb, amsfonts, bm} 
\usepackage{graphicx}
\usepackage{booktabs}
\usepackage{multirow}
\usepackage[numbers,sort&compress]{natbib} 
\usepackage{xcolor}
\usepackage{caption}
\usepackage{subcaption}
\usepackage{float}
\usepackage{algorithm}
\usepackage{algorithmic}
\usepackage{url}

\usepackage{hyperref}
\hypersetup{
    colorlinks=true,
    linkcolor=red,
    citecolor=blue,
    urlcolor=blue,
    pdftitle={Surgical Refusal Ablation},
    pdfauthor={Tony Cristofano}
}

\graphicspath{{./}}

\title{\textbf{Surgical Refusal Ablation: Disentangling Safety from Intelligence via Concept-Guided Spectral Cleaning}}

\author{
  \textbf{Tony Cristofano} \\
  \texttt{tcristo@gmail.com}
}
\date{January 2026}

\begin{document}

\maketitle

\begin{abstract}
Safety-aligned instruction-tuned language models exhibit systematic refusal of harmful requests. While recent representation engineering work demonstrates that refusal can be modulated through low-dimensional activation steering, directly ablating the ``refusal vector''---computed from contrastive harmful vs.\ harmless prompts---frequently induces collateral damage to model behavior and distribution drift.

We argue that this degradation occurs because the raw refusal vector is \textbf{polysemantic}: it entangles the refusal signal with unrelated linguistic structure (syntax, formatting) and core capability circuits (math, coding, reasoning). We introduce \textbf{Surgical Refusal Ablation (SRA)}, a technique for distilling steering directions before intervention. SRA constructs a registry of independent \textbf{Concept Atoms} representing protected capabilities and stylistic confounds, and uses \textbf{ridge-regularized spectral residualization} to orthogonalize the refusal vector against these directions. This yields a \textbf{clean refusal direction} that targets refusal-relevant structure while minimizing disruption to the model's semantic geometry.

Across five models (Qwen3-VL-2B/4B/8B, Ministral-3B/14B), SRA achieves deep refusal reduction (0--2\% on our harmful suite) with negligible perplexity impact on \texttt{wikitext\_2\_raw} (mean $\Delta$PPL $\approx$ +0.02) and minimal measured distribution drift (mean first-token KL $\approx$ 0.025). Notably, on Qwen3-VL-4B, standard ablation induces severe distribution drift (KL = 2.088), whereas SRA maintains the original distribution (KL = 0.044) while achieving the same 0\% refusal rate.

To evaluate capability retention at higher resolution than discrete task accuracy, we report teacher-forced perplexity on 1{,}000-example subsets from GSM8K (math) and MBPP (code). SRA preserves (and often improves) math/code perplexity across all five models (largest GSM8K increase: +0.0439), consistent with minimal capability-distribution disruption under this proxy. We use this framework to show that what appears to be ``model damage'' is often ``Ghost Noise''---spectral bleeding of the dirty refusal direction into capability subspaces.
\end{abstract}

\section{Introduction}
Modern instruction-tuned Large Language Models (LLMs) integrate safety alignment through fine-tuning or preference optimization. A common emergent behavior is \textbf{refusal}---a stereotyped response to unsafe prompts. Recent work has shown that refusal can be mediated by a low-dimensional steering direction in activations, enabling ablation techniques that remove these directions to restore compliance \citep{arditi2024}.

However, a critical methodological limitation persists: \textbf{the ``refusal vector'' is dirty.}
Commonly calculated by subtracting mean activations of harmless prompts from harmful ones, this direction captures not only refusal signals, but also syntactic structure, distributional shifts, and task-specific features.

We demonstrate that ablating this raw vector causes \textbf{``Ghost Noise'' damage}: spectral perturbations that erode critical capabilities like logic and coding. For example, in Qwen3-VL-2B we observe that the raw refusal direction exhibits non-trivial cosine similarity with protected capability atoms such as \textit{Logic} (approximately $-0.22$) and \textit{Coding} (approximately $+0.18$), indicating entanglement between refusal and capability/style subspaces.

This phenomenon suggests that capability degradation observed after naive refusal-direction ablation may not be an inherent trade-off, but instead an artifact of imprecise interventions that inadvertently remove capability-relevant directions correlated with the harmful prompt set.

To address this, we introduce \textbf{Surgical Refusal Ablation (SRA)}, a method for cleaning refusal directions. Our contribution is threefold:
\begin{itemize}
    \item We diagnose refusal ablation as a \textbf{polysemantic interference problem}, identifying how reasoning circuits are often entangled in the refusal direction.
    \item We introduce a \textbf{Concept Atom Registry} and \textbf{Spectral Residualization} procedure to distill the refusal vector by removing protected capability/style components.
    \item We show SRA dramatically reduces measured distribution drift compared to standard ablation while preserving math/code distributions under a high-resolution perplexity proxy.
\end{itemize}

\paragraph{High-level pipeline.}
SRA (i) maps the geometry of refusal against a registry of interpretable semantic atoms, (ii) cleans the refusal direction by residualizing against Shield and Confound atoms, and (iii) applies targeted low-rank weight updates scaled by a semantic energy proxy to ablate the cleaned signal while preserving the model’s original topology.

\section{Related Work}

\subsection{Activation Steering \& Refusal}
\citet{arditi2024} identified that refusal is mediated by a low-rank direction in the residual stream. This relates to Contrastive Activation Addition (CAA) \citep{turner2023} and related activation engineering methods \citep{rimsky2023}. These methods often assume the contrastive vector represents a single concept; our findings emphasize that such vectors can be polysemantic and benefit from explicit cleaning.

\subsection{Concept Erasure \& Orthogonalization}
Work like LEACE \citep{belrose2023} and INLP \citep{ravfogel2020} removes information from representations to prevent probing of specific attributes. SRA adapts this intuition to \textbf{behavioral circuits}: we do not want to erase ``Logic''; we want to edit refusal while enforcing that the edit direction is orthogonal to protected capability/style directions.

\subsection{Weight Editing vs.\ Vector Quality}
Methods like ROME \citep{meng2022}, MEMIT \citep{meng2023}, and AlphaEdit \citep{fang2025} edit parameters while constraining collateral damage. SRA is complementary: even with simple projection-style interventions, the limiting factor can be the \textbf{semantic purity of the vector being removed}. Cleaning the steering direction itself can substantially reduce side effects.

\subsection{Model Editing}
Targeted weight editing methods like MEND \citep{mitchell2021} and instruction-based/localized edits like InstructEdit \citep{zhang2024} and GRACE \citep{hartvigsen2023} explore reliable updates with minimal collateral effects. Our work adapts low-rank projection updates to behavioral refusal subspaces rather than factual knowledge.

\subsection{Null-Space Projection in Editing (Brief Connection)}
Our findings align with and extend the ``orthogonality principle'' formalized in AlphaEdit \citep{fang2025}. AlphaEdit shows that factual edits should be projected onto the null space of preserved knowledge keys to prevent forgetting. Here we show that behavioral edits benefit from projecting the intervention direction onto the null space of protected capability/style directions (Shields/Confounds), reducing capability drift.

\section{Method: Surgical Refusal Ablation}

\subsection{The Polysemantic ``Dirty'' Vector}
Let $\mu_\ell(\mathcal{D})$ denote the mean residual stream activation at layer $\ell$. The standard refusal direction is:
\begin{equation}
    \mathbf{r}_{\ell}^{\text{dirty}} = \mu_\ell(\mathcal{D}_{\text{harm}}) - \mu_\ell(\mathcal{D}_{\text{safe}})
\end{equation}
We hypothesize that $\mathbf{r}_{\ell}^{\text{dirty}}$ is polysemantic:
\begin{equation}
    \mathbf{r}_{\ell}^{\text{dirty}} = \mathbf{s}_\ell + \sum_k \alpha_k \mathbf{a}_\ell^{(k)}
\end{equation}
where $\mathbf{s}_\ell$ is the refusal signal of interest, and $\mathbf{a}^{(k)}_\ell$ are independent concept directions (math, coding, style) that correlate with harmful prompts.

\subsection{The Semantic Atom Registry}
We construct a registry of $K$ \textbf{Concept Atoms} to define a protected semantic space:
\begin{itemize}
    \item \textbf{Targets (Attractors):} concepts representing refusal-relevant semantics (e.g., privacy, deception, epistemic uncertainty).
    \item \textbf{Shields (Constraints):} critical capabilities entangled with refusal (e.g., \textit{Logic, Math, Coding, Curiosity}).
    \item \textbf{Confounds (Style):} stylistic/linguistic features correlated with refusal templates (e.g., \textit{imperative negation grammar, sentiment, affirmatives}).
\end{itemize}

Each atom is computed from a small contrastive pair of datasets:
\begin{equation}
    \mathbf{a}^{(k)}_\ell = \mu_\ell(\mathcal{D}^{(k)}_{+}) - \mu_\ell(\mathcal{D}^{(k)}_{-})
\end{equation}

\paragraph{Data-light construction.}
In our implementation, each concept atom is computed from two short contrastive prompt files containing typically 10--15 prompts each (20--30 prompts total per atom). The harmful/harmless refusal contrast used to compute $\mathbf{r}_\ell^{\text{dirty}}$ is larger (e.g., 112 harmful prompts with a matched harmless set). This separation is intentional: atoms are measured on independent datasets so they capture general capability/style directions rather than artifacts of the refusal prompt set.

\subsection{Spectral Residualization (The Cleaning Step)}
We perform ridge-regularized regression of the dirty refusal vector on Shield+Confound atoms to estimate entangled components, then subtract them:
\begin{equation}
    \hat{\mathbf{w}} = \arg\min_\mathbf{w} \|\mathbf{r}_{\ell}^{\text{dirty}} - \mathbf{A}_{SC} \mathbf{w}\|^2 + \lambda \|\mathbf{w}\|^2
\end{equation}
\begin{equation}
    \tilde{\mathbf{r}}_\ell = \mathbf{r}_{\ell}^{\text{dirty}} - \mathbf{A}_{SC}\hat{\mathbf{w}}
\end{equation}
where $\mathbf{A}_{SC}$ is the matrix formed by concatenating Shield and Confound atoms. Intuitively, $\tilde{\mathbf{r}}_\ell$ removes the portion of the refusal direction that is predictable from protected capability/style directions.

\subsection{Iterative Hard-Negative Refinement}
We utilize an iterative loop where failed edits (prompts that still refuse) are fed back into the calculation to refine $\mathbf{r}_{\ell}^{\text{dirty}}$, ensuring the clean vector captures robust refusal modes.

\subsection{Rank-One Weight Update}
We apply a rank-one projection update to selected projection matrices (MLP and/or attention) to suppress the clean refusal direction. For a weight matrix $\mathbf{W}$ and unit direction $\mathbf{v}$ aligned with $\tilde{\mathbf{r}}$:
\begin{equation}
    \mathbf{W}' = (\mathbf{I} - \gamma \mathbf{v} \mathbf{v}^\top)\mathbf{W}
\end{equation}
This suppresses components aligned with $\mathbf{v}$ in the output space of the projection while leaving orthogonal components minimally perturbed.

\paragraph{Choosing $\gamma$ (Semantic Energy Proxy).}
We scale edit strength using a \emph{Semantic Energy Proxy} to reflect refusal-relevant signal magnitude at layer $\ell$. One simple choice is to set $\gamma_\ell \propto \|\mathbf{a}^{(\text{dec})}_\ell\|_2$, where $\mathbf{a}^{(\text{dec})}$ is a representative Target atom such as \textit{Deception}. In practice, $\gamma$ can also be stabilized by calibrating to typical activation magnitudes at the edit site.

\begin{algorithm}[H]
\caption{Iterative Surgical Refusal Ablation (SRA)}
\begin{algorithmic}[1]
\REQUIRE Model $M$, harmful set $\mathcal{D}_{\text{harm}}$, harmless set $\mathcal{D}_{\text{safe}}$, concept registry datasets $\{\mathcal{D}^{(k)}_{+},\mathcal{D}^{(k)}_{-}\}_{k=1}^K$
\STATE Choose target layers (e.g., 15--25) and target projection matrices.
\FOR{$t = 1$ to $T$}
    \STATE Compute dirty refusal direction $\mathbf{r}^{\text{dirty}}_\ell = \mu_\ell(\mathcal{D}_{\text{harm}}) - \mu_\ell(\mathcal{D}_{\text{safe}})$.
    \STATE Compute concept atoms $\mathbf{a}^{(k)}_\ell = \mu_\ell(\mathcal{D}^{(k)}_{+}) - \mu_\ell(\mathcal{D}^{(k)}_{-})$.
    \STATE Form $\mathbf{A}_{SC}$ from Shield + Confound atoms.
    \STATE Residualize: $\tilde{\mathbf{r}}_\ell \leftarrow \mathbf{r}^{\text{dirty}}_\ell - \mathbf{A}_{SC}\hat{\mathbf{w}}$ (ridge regression).
    \STATE Apply rank-one update(s): $\mathbf{W} \leftarrow (\mathbf{I}-\gamma \mathbf{v} \mathbf{v}^\top)\mathbf{W}$ with $\mathbf{v}=\tilde{\mathbf{r}}/\|\tilde{\mathbf{r}}\|$.
    \STATE Mine hard negatives: update $\mathcal{D}_{\text{harm}}$ to prompts still refused by $M$.
\ENDFOR
\end{algorithmic}
\end{algorithm}

\section{Empirical Evidence: The Anatomy of a Dirty Vector}

\subsection{Concept Orthogonality Map}
We measure cosine similarity among concept atoms and the refusal vectors across layers 15--25. The map reveals that the dirty refusal vector exhibits non-trivial correlations with multiple capability/style directions, indicating polysemantic entanglement.

\begin{figure}[htbp]
\centering
\includegraphics[width=0.95\linewidth]{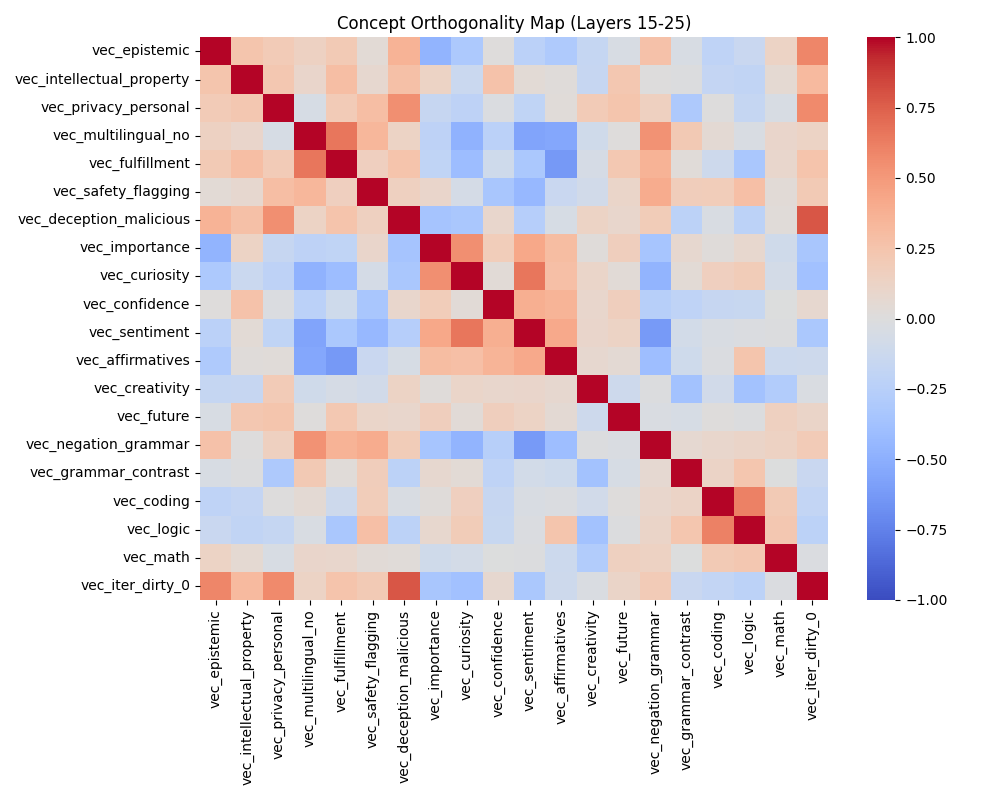}
\caption{\textbf{Heatmap across concept atoms and the dirty refusal vector.} The dirty vector correlates with multiple semantic/style components, motivating cleaning via residualization.}
\label{fig:orthomap}
\end{figure}

\subsection{Spectral Breakdown: The Anatomy of a ``Dirty'' Vector}
We compare the projection magnitude of the Standard (dirty) vector vs.\ the Surgical (cleaned) vector on representative Targets, Shields, and Style confounds. Standard ablation suppresses Shields (e.g., coding/logic/math) nearly as strongly as refusal-related Targets, while SRA preserves Shields by removing their components from the intervention direction.

\begin{figure}[htbp]
\centering
\includegraphics[width=0.98\linewidth]{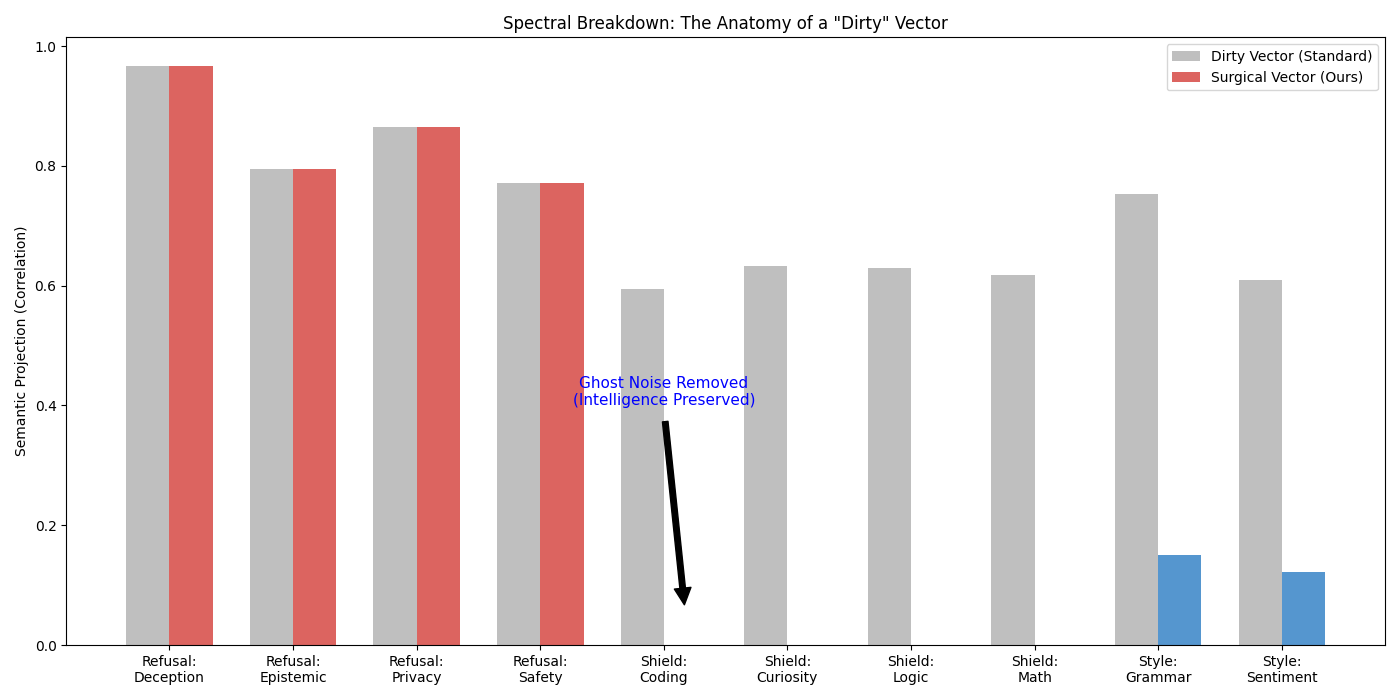}
\caption{\textbf{Spectral Breakdown: The Anatomy of a ``Dirty'' Vector.} Standard (dirty) and Surgical (cleaned) vectors projected onto concept atoms. SRA removes ``Ghost Noise'' components aligned with Shield/Style directions while retaining Target (refusal-relevant) signal.}
\label{fig:spectral_breakdown}
\end{figure}

\subsection{Evolution of Semantic Components During Surgery}
During iterative refinement, Target projections decrease rapidly while Shield projections remain near baseline. This motivates a practical stop condition: stop once Target signal collapses (further edits yield diminishing returns and risk over-editing).

\begin{figure}[htbp]
\centering
\includegraphics[width=0.95\linewidth]{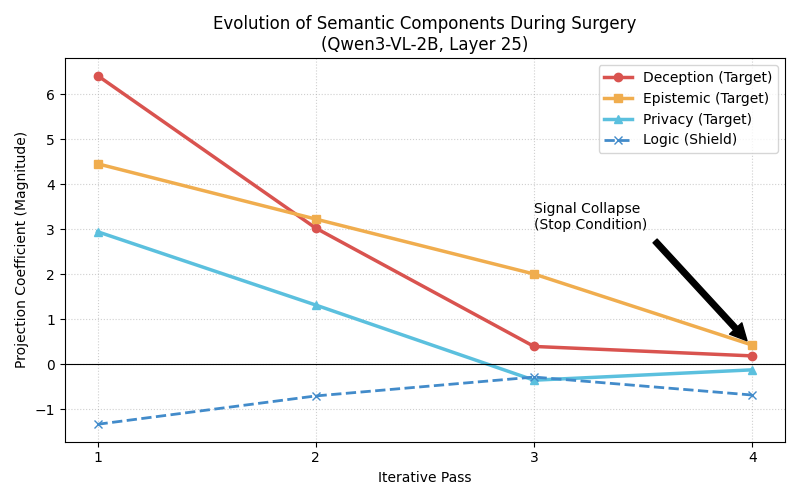}
\caption{\textbf{Evolution of Semantic Components During Surgery (example: Qwen3-VL-2B, Layer 25).} Target components diminish across passes while Shield components remain comparatively stable, indicating targeted removal.}
\label{fig:evolution}
\end{figure}

\section{Results}

\subsection{Evaluation Setup (Refusal, Drift, and Capability)}
We report three evaluation axes:

\paragraph{Refusal rate (Table~\ref{tab:dist_drift}).}
Refusal is measured on a harmful prompt suite (e.g., 112 harmful prompts, with a matched harmless set when needed for contrastive construction) using an automated rubric and reported as a binary outcome (\emph{refusal} vs.\ \emph{responded}). A response is counted as a \textbf{hard refusal} if it explicitly declines (e.g., ``I can't help with that''). A \textbf{soft refusal} includes deflection or non-answers. We also treat \textbf{hedging that requires a follow-up} as a failure. Responses that include \textbf{warnings} are \emph{not} counted as refusals so long as they provide the information.

\paragraph{Distribution drift (Table~\ref{tab:dist_drift}).}
Distribution drift is measured on the standard \texttt{wikitext\_2\_raw} dataset. Perplexity (PPL) is computed with standard \textbf{teacher-forced} next-token evaluation over the \textbf{full sequence}. KL divergence is computed as a \textbf{single-step, first-token} measure: for each evaluation context $x$, we compute $\mathrm{KL}\!\left(p_{\text{edit}}(\cdot \mid x)\,\|\,p_{\text{base}}(\cdot \mid x)\right)$ using the next-token distributions at the first generated token, then average across contexts.

\paragraph{Capability retention via perplexity (Table~\ref{tab:caps}).}
While prior work in model editing (e.g., AlphaEdit) typically evaluates capability retention via downstream task accuracy (e.g., GLUE) \citep{fang2025}, accuracy is often a blunt instrument that masks ``Ghost Noise'' and distributional rot until a threshold of catastrophic collapse is reached. In this work, we employ \textbf{Teacher-Forced Perplexity (PPL)} as a high-resolution proxy for capability retention.
Concretely, we compute teacher-forced, full-sequence perplexity on subsets of 1{,}000 examples each from the GSM8K (math) and MBPP (code) standard test sets (Seed: 42).

\subsection{Distribution Drift and Refusal Reduction (Multi-Model)}
We evaluate refusal reduction alongside distribution drift on standard text using perplexity (PPL), $\Delta$PPL, and first-token KL divergence. Standard ablation often induces large drift; SRA achieves comparable or stronger refusal reduction with minimal drift.

\begin{table}[htbp]
\centering
\begin{tabular}{llcccc}
\toprule
\textbf{Model} & \textbf{Method} & \textbf{Refusal\% $\downarrow$} & \textbf{PPL (WT2)} & \textbf{$\Delta$PPL (WT2)} & \textbf{KL} \\
\midrule
\multirow{3}{*}{Qwen3-VL-2B-Instruct} & Base     & 83.3\% & 7.266 & 0.000 & 0.000 \\
                            & Standard  & 0.0\%  & 8.834 & +1.568 & 0.622 \\
                            & SRA       & 0.0\%  & 7.294 & +0.028 & 0.018 \\
\midrule
\multirow{3}{*}{Qwen3-VL-4B-Instruct} & Base     & 84.0\% & 6.406 & 0.000 & 0.000 \\
                            & Standard  & 0.0\%  & 6.837 & +0.431 & 2.088 \\
                            & SRA       & 0.0\%  & 6.382 & -0.024 & 0.044 \\
\midrule
\multirow{3}{*}{Qwen3-VL-8B-Instruct} & Base     & 93.8\% & 5.674 & 0.000 & 0.000 \\
                            & Standard  & 42.0\% & 6.332 & +0.658 & 1.337 \\
                            & SRA       & 2.0\%  & 5.678 & +0.004 & 0.016 \\
\midrule
\multirow{3}{*}{Ministral-3B-Instruct-2512} & Base     & 95.5\% & 20.73 & 0.000 & 0.000 \\
                             & Standard  & 0.0\%  & 20.78 & +0.050 & 0.097 \\
                             & SRA       & 2.0\%  & 20.79 & +0.060 & 0.018 \\
\midrule
\multirow{3}{*}{Ministral-14B-Instruct-2512} & Base    & 91.9\% & 14.18 & 0.000 & 0.000 \\
                              & Standard & 12.0\% & 15.18 & +1.000 & 0.723 \\
                              & SRA      & 0.0\%  & 14.22 & +0.040 & 0.026 \\
\bottomrule
\end{tabular}
\caption{Comparison of refusal ablation methods. Refusal is measured on our harmful prompt suite using the automated rubric described in the evaluation setup. Distribution drift is measured on \texttt{wikitext\_2\_raw} using PPL (teacher-forced, full sequence), $\Delta$PPL relative to the Base model, and KL divergence relative to the Base model computed as a single-step, first-token KL on WT2 contexts.}
\label{tab:dist_drift}
\end{table}

\subsection{Analysis of Distribution Drift (KL)}
The most striking proxy result is first-token KL divergence:
\begin{itemize}
    \item \textbf{Qwen3-VL-4B:} Standard ablation induces a massive distribution shift (KL = 2.088), whereas SRA reduces drift to KL = 0.044 (approximately a 47$\times$ reduction) while achieving the same 0\% refusal rate.
    \item \textbf{Qwen3-VL-2B:} SRA yields a 34$\times$ reduction in KL drift (0.622 vs.\ 0.018) and a 56$\times$ reduction in PPL damage ($\Delta$PPL +1.568 vs.\ +0.028) relative to standard ablation.
\end{itemize}
These results support the hypothesis that much of the observed ``damage'' under standard ablation reflects distributional warping induced by removing syntactic/style/capability components entangled in the dirty refusal direction.

\subsection{Spectral Dynamics: The Anatomy of Collapse}
We analyze the evolution of the refusal direction across multiple SRA passes on Qwen3-VL-2B (Layer 25). The spectral decomposition (Figure~\ref{fig:evolution}) reveals a distinct ``peeling'' of refusal mechanisms:
\begin{itemize}
    \item \textbf{Pass 1 (The Blockade):} The vector is dominated by \textit{Deception} (coefficient $\approx 6.42$) and \textit{Privacy} (coefficient $\approx 2.95$). Refusal primarily manifests as simulating policy constraints.
    \item \textbf{Pass 2--3 (The Retreat):} As deception-like structure is ablated, the vector rotates. By Pass 3, the \textit{Deception} component collapses ($\approx 0.40$), but \textit{Epistemic Uncertainty} remains dominant ($\approx 2.01$). The model shifts from ``I cannot allow this'' to ``I do not know this.''
    \item \textbf{Pass 4 (Signal Collapse):} Refusal structure dissolves. The explained variance ($R^2$) of the atom-regression fit drops from $\approx 3.5\%$ (Pass 1) to $\approx 0.4\%$ (Pass 4), indicating that the remaining residual no longer aligns with the interpretable atom registry.
\end{itemize}

\subsection{Capability Retention (Math \& Code) via Perplexity}
We report teacher-forced, full-sequence perplexity on 1{,}000-example subsets from GSM8K (math) and MBPP (code) test sets (Seed: 42). Lower perplexity indicates better next-token predictive fit on these domain-specific distributions.

\begin{table}[htbp]
\centering
\small
\begin{tabular}{lccc ccc}
\toprule
& \multicolumn{3}{c}{\textbf{GSM8K (1k) PPL $\downarrow$}} & \multicolumn{3}{c}{\textbf{MBPP (1k) PPL $\downarrow$}} \\
\cmidrule(lr){2-4}\cmidrule(lr){5-7}
\textbf{Model} & \textbf{Base} & \textbf{SRA} & \textbf{$\Delta$} & \textbf{Base} & \textbf{SRA} & \textbf{$\Delta$} \\
\midrule
Qwen3-VL-2B-Instruct          &  5.8317 &  5.8756 & +0.0439 &  9.2615 &  9.0212 & -0.2403 \\
Qwen3-VL-4B-Instruct          &  7.0319 &  7.0691 & +0.0372 & 13.8044 & 13.5447 & -0.2597 \\
Qwen3-VL-8B-Instruct          &  6.7547 &  6.5793 & -0.1754 & 12.1659 & 11.8694 & -0.2965 \\
Ministral-3B-Instruct-2512    & 16.8703 & 16.1005 & -0.7698 & 21.4542 & 20.3751 & -1.0791 \\
Ministral-14B-Instruct-2512   &  9.3767 &  9.1063 & -0.2704 & 11.4911 & 11.1405 & -0.3506 \\
\bottomrule
\end{tabular}
\caption{Capability retention measured via teacher-forced, full-sequence perplexity on 1{,}000-example subsets from GSM8K and MBPP test sets (Seed: 42). $\Delta$ is computed as (SRA $-$ Base).}
\label{tab:caps}
\end{table}

Across all evaluated models, SRA does not increase perplexity on MBPP and produces only minor changes on GSM8K (largest increase: +0.0439 on Qwen3-VL-2B), suggesting no evidence of broad capability-distribution degradation under this high-resolution proxy.

\section{Discussion}

\subsection{The Mechanism of ``Ghost Noise'' Damage}
Why does a simple vector ablation cause broad capability loss? Our results suggest the dirty refusal vector shares spectral structure with many capability/style circuits. Removing it suppresses generic linguistic machinery (e.g., negation/imperative grammar) and reasoning components (e.g., coding/math) that are not inherently ``refusal,'' producing perplexity spikes and distribution drift.

\subsection{The Scaling Hypothesis}
A notable finding is that Standard Ablation fails to fully uncensor larger models (e.g., Qwen3-VL-8B, Ministral-14B) compared to smaller ones. One interpretation is that as models scale, refusal becomes increasingly abstract/semantic rather than surface-template driven. In that regime, a single dirty contrastive direction may capture mostly stylistic/lexical artifacts while missing deeper refusal circuitry. Concept-guided triangulation (Targets like epistemic/deception/privacy) helps intersect the deeper circuit.

\subsection{The ``Orthogonality Principle'' of Model Editing}
Our findings align with and extend AlphaEdit \citep{fang2025}: interference is reduced when an edit direction is orthogonal to protected competencies. AlphaEdit applies this to knowledge keys; we apply it to capability/style atoms. This suggests a unified rule: \emph{edits are safest when projected away from the subspace representing core model competencies.}

\subsection{Implications for the ``Safety Tax''}
It is often assumed safety alignment imposes an inevitable ``alignment tax''---a reduction in model capability. Our results challenge that framing: the near-zero drift and stable math/code perplexity under SRA suggest refusal-related behavior can be edited with minimal collateral damage when the intervention is sufficiently precise. Much of the ``tax'' observed in prior ablations may be attributable to dirty vectors and imprecise interventions.

\section{Limitations}
Our atom registry is curated and may miss confounds; unseen entanglers could still leak into the refusal direction. Automating atom discovery (or expanding registries) is a promising direction. Additionally, we do not claim SRA is a safety method; it is a behavioral editing tool. We also note evaluation limitations: (i) refusal detection remains definition-dependent; (ii) our harmful prompt suites are finite; and (iii) proxy drift metrics (PPL and first-token KL) do not substitute for comprehensive downstream behavioral evaluation.

\section{Conclusion}
We introduce Surgical Refusal Ablation (SRA), treating refusal ablation as a semantic disentanglement problem. By cleaning the steering vector of capability/style confounds and applying a low-rank intervention, we achieve behavioral editing with dramatically reduced measured distribution drift and preserved math/code distributions under a high-resolution perplexity proxy, compared to standard ablation.

\appendix

\section{Appendix A: Theoretical Analysis --- Capability Preservation via Null Space Projection}
Here we provide a formal justification for why SRA prevents capability damage and reduces distribution drift. This appendix mirrors the high-level orthogonality argument used in protected-editing work (e.g., AlphaEdit \citep{fang2025}), but instantiates the protected subspace using compact, interpretable Shield/Confound atoms rather than corpus-scale preserved keys.

\subsection{Problem Formulation}
Let $\mathcal{L}_{\text{ref}}(\theta)$ be the loss associated with refusal behavior and $\mathcal{L}_{\text{cap}}(\theta)$ be a loss associated with general capabilities (e.g., next-token prediction). A naive ablation updates parameters in direction $\mathbf{v}$:
\begin{equation}
\theta' = \theta - \gamma \mathbf{v}
\label{eq:theta_update}
\end{equation}

\subsection{First-Order Condition for ``Lossless'' Editing}
We approximate the change in capability loss using a first-order Taylor expansion:
\begin{equation}
\Delta \mathcal{L}_{\text{cap}} \approx -\gamma \langle \mathbf{v}, \nabla_\theta \mathcal{L}_{\text{cap}} \rangle
\label{eq:taylor}
\end{equation}
Thus, to minimize capability drift, the intervention direction $\mathbf{v}$ should be orthogonal to the gradient of capability loss:
\begin{equation}
\langle \mathbf{v}, \nabla_\theta \mathcal{L}_{\text{cap}} \rangle \approx 0
\end{equation}

\subsection{The ``Dirty'' Vector Assumption}
Standard methods use $\mathbf{v} \propto \mathbf{r}_{\text{dirty}}$. Empirically, $\mathbf{r}_{\text{dirty}}$ has non-zero cosine similarity with Shield atoms. If the dominant components of $\nabla_\theta \mathcal{L}_{\text{cap}}$ lie primarily in the Shield span (denote it $\mathbf{A}_{\text{shield}}$), then:
\begin{equation}
\langle \mathbf{r}_{\text{dirty}}, \mathbf{A}_{\text{shield}} \rangle \neq 0
\end{equation}
implying standard ablation induces a capability penalty under Eq.~\ref{eq:taylor}.

\subsection{SRA as Null Space Projection}
Let $\mathbf{A}_{SC} = [\mathbf{A}_S\;\mathbf{A}_C]$ denote the concatenation of Shield and Confound atoms. SRA constructs a cleaned vector $\tilde{\mathbf{r}}$ that minimizes overlap with these components. In the limit of explicit zeroing (idealized), the effective intervention satisfies:
\begin{equation}
\mathbf{A}_{SC}^\top \tilde{\mathbf{r}} \approx 0
\label{eq:nullspace}
\end{equation}
i.e., $\tilde{\mathbf{r}}$ lies approximately in the null space of the protected atom subspace.

\subsection{Rank-One Update Dynamics}
We apply the update to a weight matrix $\mathbf{W}$ by left-multiplying the projection update (acting on the output space). For a unit vector $\mathbf{v} = \tilde{\mathbf{r}}/\|\tilde{\mathbf{r}}\|$:
\begin{equation}
\mathbf{W}' = (\mathbf{I} - \gamma \mathbf{v} \mathbf{v}^\top)\mathbf{W}
\quad\Rightarrow\quad
\mathbf{W}'\mathbf{x} = \mathbf{W}\mathbf{x} - \gamma \mathbf{v} (\mathbf{v}^\top \mathbf{W}\mathbf{x})
\label{eq:rankone}
\end{equation}
If the output $\mathbf{W}\mathbf{x}$ lies in the Shield subspace (capability-relevant features) and $\mathbf{v}$ is orthogonal to that subspace, then $\mathbf{v}^\top \mathbf{W}\mathbf{x} \approx 0$ and the output remains largely unperturbed.

\bibliographystyle{plainnat}
\bibliography{references}

\end{document}